\def\year{2018}\relax
\documentclass[letterpaper]{article} %DO NOT CHANGE THIS
\usepackage{aaai18}  %Required
\usepackage{times}  %Required
\usepackage{helvet}  %Required
\usepackage{courier}  %Required
\usepackage{url}  %Required
\usepackage{graphicx}  %Required
\usepackage{amsmath}
\usepackage{amssymb}
\usepackage{amsthm}
\usepackage{bm}
\usepackage{algorithmic}
\usepackage{algorithm}
\usepackage{mathletters}

\usepackage{subcaption}

\begin{document}
\title{Learning to Become an Expert:\\Deep Networks Applied To Super-Resolution Microscopy}
\author{Louis-\'Emile Robitaille\\
LVSN, Universit\'e Laval, Canada\\
\And
Audrey Durand\\
LVSN, Universit\'e Laval, Canada\\
\And
Marc-Andr\'e Gardner\\
LVSN, Universit\'e Laval, Canada\\
\AND
Christian Gagn\'e\\
LVSN, Universit\'e Laval, Canada\\
\And
Paul De Koninck\\
CERVO, Universit\'e Laval, Canada\\
\And
Flavie Lavoie-Cardinal\\
CERVO, Universit\'e Laval, Canada\\
}
\maketitle

\begin{abstract}
    With super-resolution optical microscopy, it is now possible to observe molecular interactions in living cells. The obtained images have a very high spatial precision but their overall quality can vary a lot depending on the structure of interest and the imaging parameters. Moreover, evaluating this quality is often difficult for non-expert users. In this work, we tackle the problem of learning the quality function of super-resolution images from scores provided by  experts. More specifically, we are proposing a system based on a deep neural network that can provide a quantitative quality measure of a STED image of neuronal structures given as input. We conduct a user study in order to evaluate the quality of the predictions of the neural network against those of a human expert. Results show the potential while highlighting some of the limits of the proposed approach. 
\end{abstract}

%%%%%%%%%%%%%%%%%%%%%%%%%%
%%%%%%%%%%%%%%%%%%%%%%%%%%
%%        INTRO         %%
%%%%%%%%%%%%%%%%%%%%%%%%%%
%%%%%%%%%%%%%%%%%%%%%%%%%%
In order to understand cellular mechanisms and their related disorders, we need to improve our knowledge of the molecular components making those cells, on their spatial dynamics, and on their signaling interactions inside subcellular compartments. These processes, occurring at the nanoscale, can now be observed in living cells thanks to recent breakthroughs in optical methods which led to the development of optical super-resolution microscopy. Among these methods, STimulated Emission Depletion (STED) microscopy~\cite{Hell1994,Klar2000,Willig2006}\footnote{Stefan W. Hell was awarded a Nobel prize in 2014 for this revolutionary microscopy technique.} overcomes the diffraction limit of light and improves the resolution of an optical microscope down to 20-25~nm (about 10 fold improvement over conventional optical microscopy).
This means that structures, such as cytoskeletal filaments or receptor clusters, closer than 200~nm from each other, cannot be distinguished on a conventional microscope but can be discerned on a STED microscope and therefore studied. With STED microscopy, we can now observe molecular structures and protein complexes of living cells in action~\cite{Sahl2017}.

\begin{figure}
    \centering
    \begin{subfigure}[b]{0.15\textwidth}
        \centering
        \includegraphics[width=\textwidth]{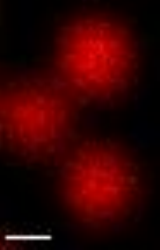}
        \caption{Confocal}
    \end{subfigure}
    \begin{subfigure}[b]{0.15\textwidth}
        \centering
        \includegraphics[width=\textwidth]{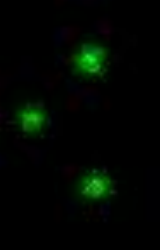}
        \caption{STED}
    \end{subfigure}
    \begin{subfigure}[b]{0.15\textwidth}
        \centering
        \includegraphics[width=\textwidth]{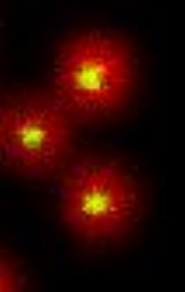}
        \caption{Merge}
    \end{subfigure}
    \caption{Fluorescence PSFs of TetraSpeck microspheres.\\Scale:~200~nm.}
\label{fig:beads}
\end{figure}

\begin{figure}
    \centering
    \begin{subfigure}[b]{0.47\textwidth}
        \centering
        \includegraphics[width=\textwidth]{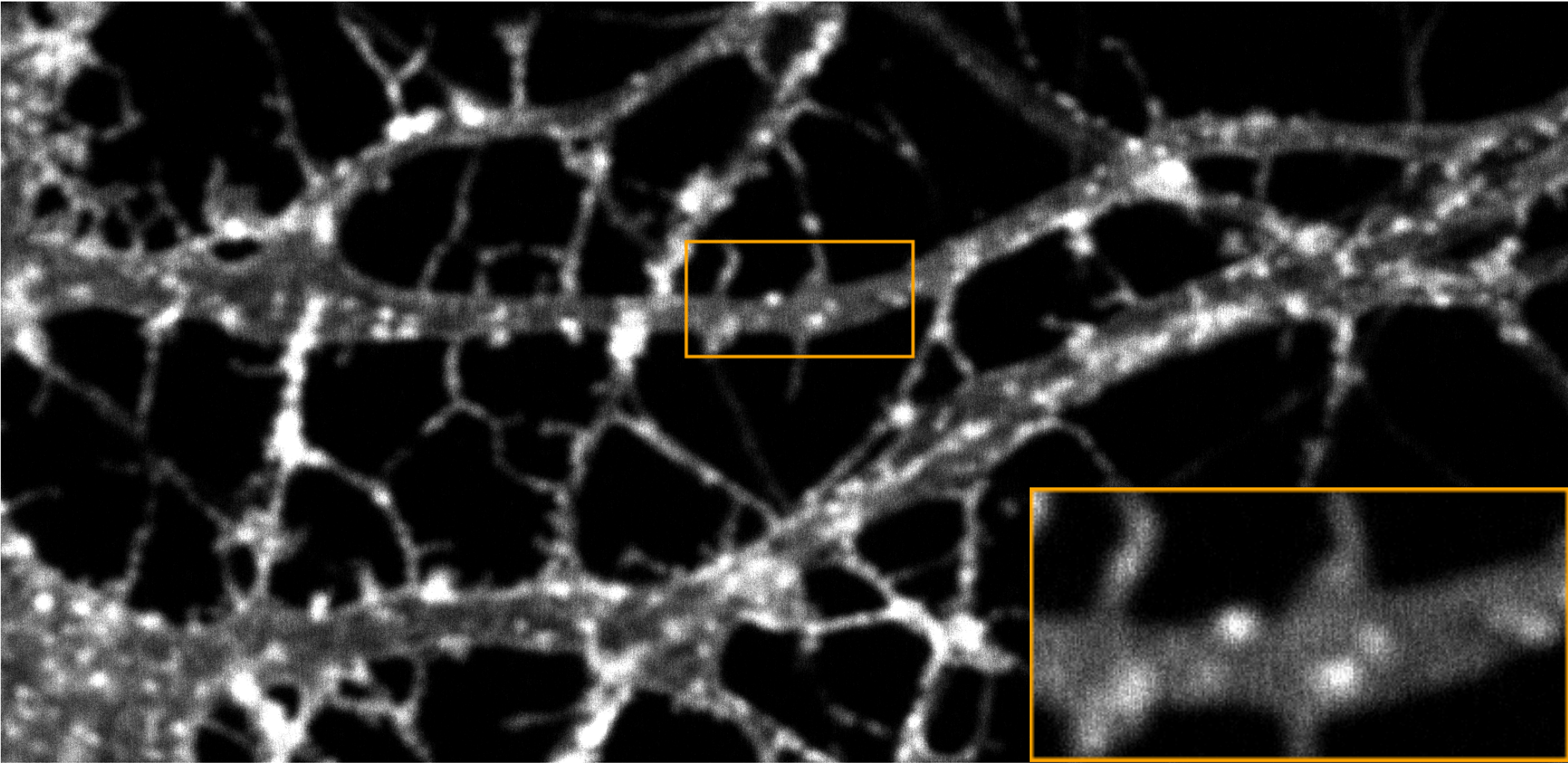}
        \caption{Confocal}
    \end{subfigure}
    
    \begin{subfigure}[b]{0.47\textwidth}
        \centering
        \includegraphics[width=\textwidth]{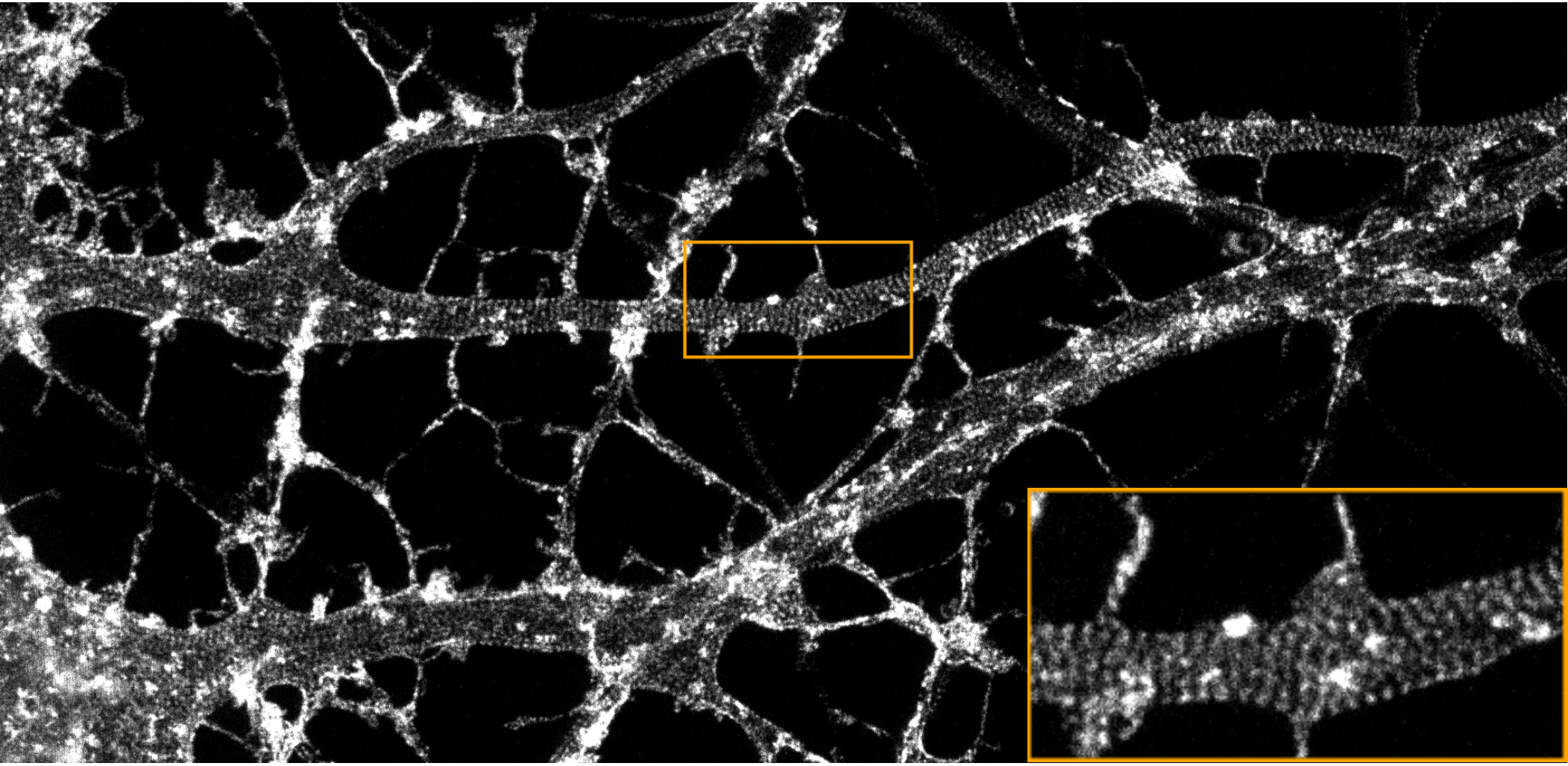}
        \caption{STED}
    \end{subfigure}
    \caption{Imaging of actin protein on fixed hippocampal neurons.}
\label{fig:neurons}
\end{figure}

Based on a confocal laser scanning microscope, a STED microscope overcomes the diffraction barrier using the combination of a Gaussian excitation beam and a donut-shaped depletion beam (deactivation beam) for selective deactivation of fluorescent markers.
Fluorophores are chemical compounds or proteins having the property of emitting light each time they are stimulated (excited) by a specific wavelength. These can be either coupled to antibodies (chemical compounds) that will recognize a protein of interest or fused directly on the protein (fluorescent proteins) and therefore make a structure visible on a fluorescence microscope.
Using the depletion beam, fluorophores can be switched from a fluorescent on-state to a non-fluorescent dark state by means of stimulated emission and, due to the donut shape of the depletion beam, only the fluorophores in the middle of the donut can still be detected. Saturation of the deactivation effectively reduces the size of the diffraction-limited fluorescent spots that is emitted by the fluorophores being imaged.
Figure~\ref{fig:beads} shows fluorescence PSFs (Point Spread Functions) obtained by confocal and STED imaging of TetraSpeck microspheres (100~nm), along with their merge. We observe that the STED image has a much higher resolution than the confocal image. Moreover, we can see the donut shape on the merged image.
Figure~\ref{fig:neurons} shows neurons images obtained by confocal and STED microscopy. We observe that STED microscopy allows us to observe a periodical lattice structure of the neuronal actin protein which is invisible with conventional confocal microscopy.

Super-resolution microscopes are highly specialized devices, significantly more complex to use than conventional optical microscopes. This reduces their accessibility, circumventing its use by a large community of scientists. Moreover, the overall quality of the obtained images can vary a lot depending on the imaging parameters or the biological structure of interest. Often, it can be very difficult to evaluate the quality of such images for non-expert users. Tuning the imaging parameters of such devices toward images of quality good enough for the biological task at hand is therefore a challenge.

In this work, we tackle the problem of learning to evaluate the quality of STED images. This could allow not only to support non-experts in their measurements, but also constitute a step toward a fully automated imaging system.
We address this problem using deep learning. To this end, a brand new dataset was built, which is used to train the network and assess its performance. We then conduct a user study to evaluate the capability of the network for fooling an expert in front of other experts. We also evaluate the capability of the network to generalize its quality prediction to STED images of a different protein. Results show the ability of the network to score the quality of images and also to generalize to images of other proteins.

%%%%%%%%%%%%%%%%%%%%%%%%%%
%%%%%%%%%%%%%%%%%%%%%%%%%%
%%     RELATED WORK     %%
%%%%%%%%%%%%%%%%%%%%%%%%%%
%%%%%%%%%%%%%%%%%%%%%%%%%%
\section{Problem Statement}

\begin{figure}
    \centering
    \begin{subfigure}[b]{0.23\textwidth}
        \centering
        \includegraphics[width=\textwidth]{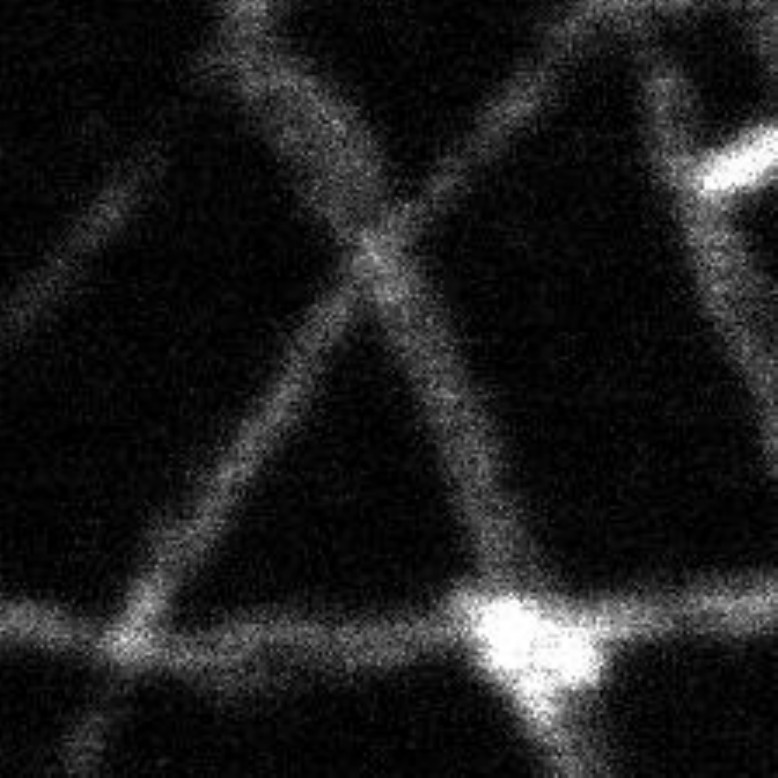}
        \caption{0.07}
    \end{subfigure}
    \begin{subfigure}[b]{0.23\textwidth}
        \centering
        \includegraphics[width=\textwidth]{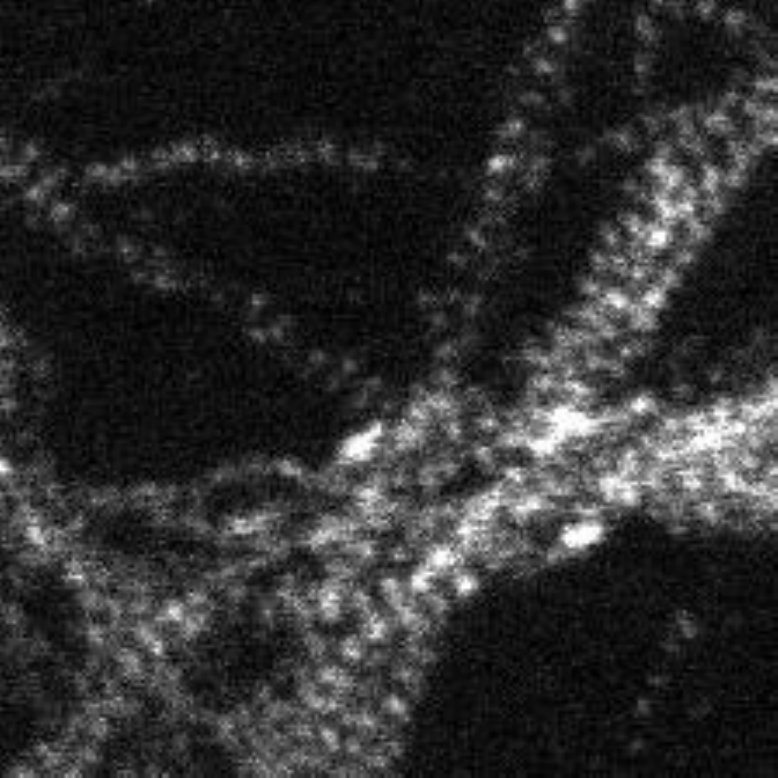}
        \caption{0.79}
    \end{subfigure}
    \caption{Examples of quality scores for two images.\\Size:~4.48~$\mu$m.}
\label{fig:data:quality_examples}
\end{figure}

Let $\cI$ denote the space of possible STED images.
We aim at learning the quality function $f: \cI \mapsto [0,1]$ that takes as input an image and outputs a quality score. This corresponds to a standard regression problem.
Figure~\ref{fig:data:quality_examples} shows examples of quality scores for two different images.
The quality score of an image incorporates several features such as the resolution of the observed structures, the signal-to-noise ratio (SNR), the deterioration of the fluorophores (photobleaching) and structure (photoxicity) due to the imaging process, or the observability of specific structures. The quality score given to an image by an expert is therefore some sort of (unknown) tradeoff between several objectives.

\subsection{Related Work}

The quality of images is typically evaluated using several metrics such as the resolution of the measured structures and the SNR. Real-time evaluation of these metrics is problematic in many situations, for example, when images are characterized by very low signals or very small structures.
While there exists methods for quantifying different aspects of a super-resolution image~\cite{Banterle2013,Durisic2014,Merino2017}, analysis of the overall quality of an image remains a challenge. Although it may be relatively simple for an expert to identify the quality of an image by looking at it, this task can become very difficult, if not impossible, for a novice user because too many aspects must be considered simultaneously.

Deep learning systems have already been applied to various tasks in microscopy imaging, such as supporting users in data analysis~\cite{Kraus2017} and image segmentation for neuron reconstruction~\cite{Li2017}.
However, these tasks did not cover the general image quality assessment problem.
The problem tackled in this work is similar to the Image Quality Assessment (IQA) problem, where one tries to learn the quality of natural images subject to noise and other artifacts (e.g., due to compression). This problem has been addressed previously with deep learning techniques~\cite{Li2011}. The difference here is that the quality concept is driven by the capability of the image to convey information toward a goal, that is the task at hand for a neuroscientist. In opposition, the quality in the typical IQA problem is driven by the aesthetic.

%%%%%%%%%%%%%%%%%%%%%%%%%%
%%%%%%%%%%%%%%%%%%%%%%%%%%
%%       APPROACH       %%
%%%%%%%%%%%%%%%%%%%%%%%%%%
%%%%%%%%%%%%%%%%%%%%%%%%%%
\section{Proposed Approach}

We propose to learn the quality function from an expert using a CNN (Convolutional Neural Network).
More specifically, we consider a network made of 6 convolutional layers and 2 fully connected layers. An ELU activation (Exponential Linear Unit) is used after each convolutional and fully connected unit. Max pooling (kernel 2x2, stride 1) is added after each convolutional unit. Batch normalization is applied to all the layers except the first one. The output is driven by a nonlinear activation (sigmoid) to retrieve a quality score between~0 and~1.

\subsection{Initial Data}

A brand new dataset has been built for the task at hand. It contains 1140 grayscale images of $224\times 224$ pixels of 20~nm.
Each image was obtained by the imaging of the protein actin marked with Phalloidin-STAR635 (Abberior, 1:50) on axons of fixed hippocampal neurons. STED imaging of the fluorophore STAR635 was performed on an Abberior 4-colors STED microscope with 40~MHz pulsed lasers at 640~nm and 775~nm for the excitation and depletion respectively.
Images of different qualities were produced by changing the acquisition parameters, that is the imaging time spent per pixel, the excitation laser voltage, and the depletion laser voltage.
Images have been labeled by an expert, where each label corresponds to a quality score in $[0,1]$.

\subsection{Training Process}

\begin{figure}
    \centering
    \includegraphics[width=0.47\textwidth]{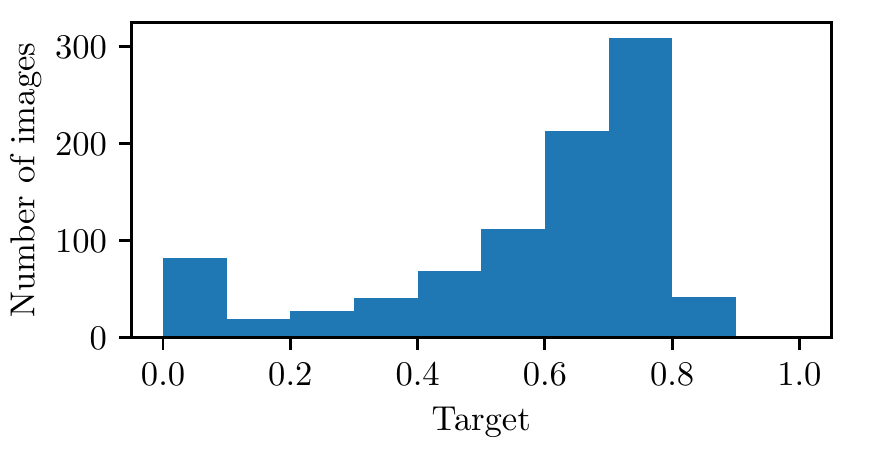}
    \caption{Distribution of the actin training set.}
\label{fig:data:actin_train_distribution}
\end{figure}

A 80/10/10 split of the randomly shuffled dataset is used for training, validation, and testing respectively. Figure~\ref{fig:data:actin_train_distribution} shows the distribution of the resulting training dataset. We observe that the dataset is not balanced, which is due to the difficulty for experts to tune the STED system toward specific imaging qualities.
However, this distribution is maintained across the splits, ensuring that the test measurement is consistent with the learning data.

\begin{figure}
    \centering
    \includegraphics[width=0.47\textwidth]{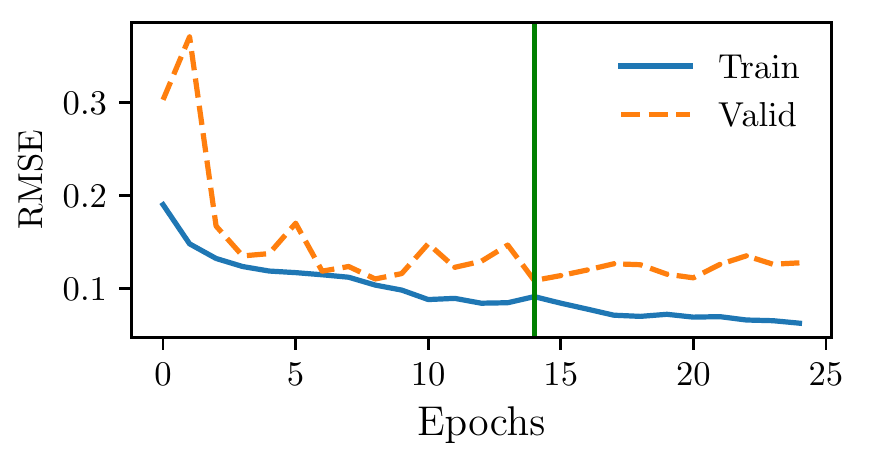}
    \caption{Learning curves.}
\label{fig:learning_curves}
\end{figure}

Data augmentation (flips and rotations) is applied on the training set, resulting in 1,824 training images.
Training is performed using stochastic gradient descent with momentum and mean squared error (MSE) as loss function. The learning rate is set to $\eta=0.01$, the momentum is set to $\alpha=0.9$, and batches of size 100 are used. Early stopping is done if no improvement has been seen on the validation dataset over 10 epochs. The weights leading to the best score on the validation dataset are kept in the final model. All the images are normalized with the mean and the standard deviation of the training dataset, here $\mu = 0.500$ and $\sigma = 2.278\times 10^{-4}$.
Images used in the test phase are also normalized using these parameters.
Figure~\ref{fig:learning_curves} shows the evolution of the root mean squared error (RMSE) on the training and validation sets. The vertical line indicates the early stopping point of the training process. 

%%%%%%%%%%%%%%%%%%%%%%%%%%
%%%%%%%%%%%%%%%%%%%%%%%%%%
%%      USER STUDY      %%
%%%%%%%%%%%%%%%%%%%%%%%%%%
%%%%%%%%%%%%%%%%%%%%%%%%%%
\section{User Study}

While using the MSE as loss function provides good results at training time, it is not very informative to assess the ability of the neural network to fully replace a human expert in the learning loop. Indeed, if experts appear to be very noisy, the MSE might be high while the system might be performing quite well.
It would therefore be interesting to compare the system predictions \emph{against} the expert, from the perspective of the expert.

We developed a web-based application that would sequentially present an expert with STED images and two scores: the target and the network prediction, in random order. The expert would either pick the most relevant score, mark both scores as \emph{equivalent}, or discard the image. The last case means that an error occured at the time of the labeling since that neither the prediction nor the target appears to be good to the expert tester. User studies of this sort for assessing the capability of a system to produce realistic results from a human perspective have been used previously, for example to evaluate whether synthetic objects look realistic when they are composited into input images~\cite{Gardner2017}.

\subsection{Measuring Performance}

Let $N$ denote the size of the test set and $\tilde N$ denote the size of the \emph{effective} test set, that is the number of images that were not discarded by the tester. Let $T$, $P$, and $E$ respectively denote the number of images where the tester picked the target, the prediction, and marked them as equivalent.

\subsubsection{Confusion}

Let the \emph{confusion} denote the situation where a human expert cannot distinguish between the prediction given by the neural network and the target given by a human expert. Confusion is explicitly indicated by the tester when two images are marked as equivalent. Confusion is also implicitly indicated by the selection of the prediction against the target. Therefore, we say that \emph{perfect confusion} is reached either if
\begin{align}
\label{eqn:perfect_confusion}
    \frac{E}{\tilde N}=1  \quad \text{or} \quad \frac{P}{T+P}= \frac{1}{2}.
\end{align}
The former equality measures the explicit confusion and is true if all effective images are marked as equivalent.
The latter equality measures the implicit confusion and is true \emph{on average} if the user picks uniformly among predictions and targets.
In order to obtain a confusion score that covers both the explicit and implicit confusion, we rewrite the implicit confusion as
\begin{align}
\label{eqn:implicit_confusion}
    2P & = T + P = \tilde N - E.
\end{align}
Hence, we have
\begin{align}
\label{eqn:confusion_first}
    \tilde N = 2P + E.
\end{align}
Equation~\ref{eqn:confusion_first} combines both conditions toward a perfect confusion score. From this, we define the confusion as the difference between $2P+E$ and it's target $\tilde N$.
Under the assumption that a system is good if it can be mistaken for a human expert by another human expert, we evaluate our technique using the following confusion measure:
\begin{align}
\label{eqn:confusion_score}
    \cC = 1-\frac{\abs{(2P + E)-\tilde N}}{\tilde N}.
\end{align}
Maximum confusion (1) is reached if the tester either selects the network prediction as much as the real target, or explicitly states the prediction and the target as equivalent.

\subsubsection{Domination}

The situation where confusion does not occur is also of interest. For example, confusion might not occur because in some situations, the network prediction may always be picked instead of the target. Thought it would be surprising, this could highlight the fact that the expert suffered from decision fatigue~\cite{Vohs2008} during the labeling process, leading to wrong labels that the network is able to correct. Concurrently with the confusion measure, we therefore consider the following domination measure:
\begin{align}
\label{eqn:domination}
    \cD = \frac{P}{T+P}.
\end{align}
Domination evaluates how much the prediction is selected compared with the target \emph{when the tester expert makes a choice}. Maximum domination (1) is reached if the tester always selects the prediction instead of the target.

\subsection{Data}

The user study is performed on the two following datasets.

\paragraph{Actin}

\begin{figure}
    \centering
    \resizebox{\columnwidth}{!}{%
    \includegraphics{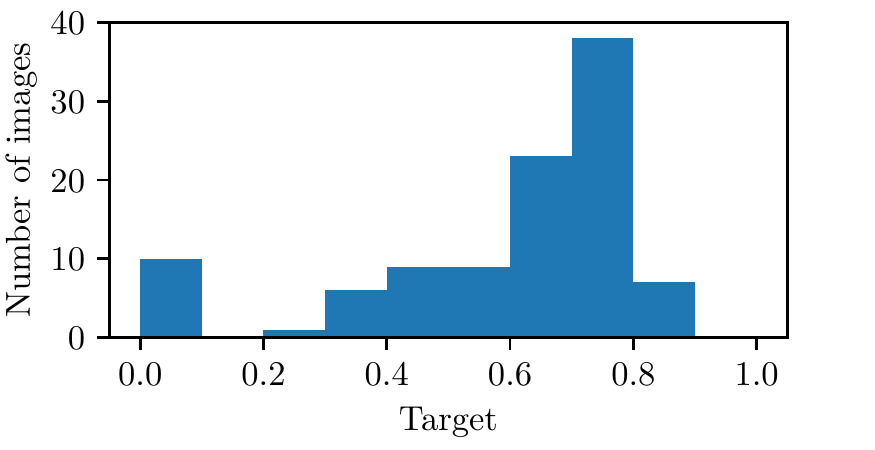}}
    \caption{Distribution of the actin testing set.}
\label{fig:data:actin_test:distribution}
\end{figure}

This dataset contains 103 images of the actin protein on fixed hippocampal neurons. More specifically, these images are taken from the 10\% test split taken from the initial data. Figure~\ref{fig:data:actin_test:distribution} shows the distribution of the resulting dataset across quality scores. We observe that it is unbalanced, but its distribution is similar to the training data (Figure~\ref{fig:data:actin_train_distribution}).

\paragraph{Tubulin}

\begin{figure}
    \centering
    \resizebox{\columnwidth}{!}{%
    \includegraphics{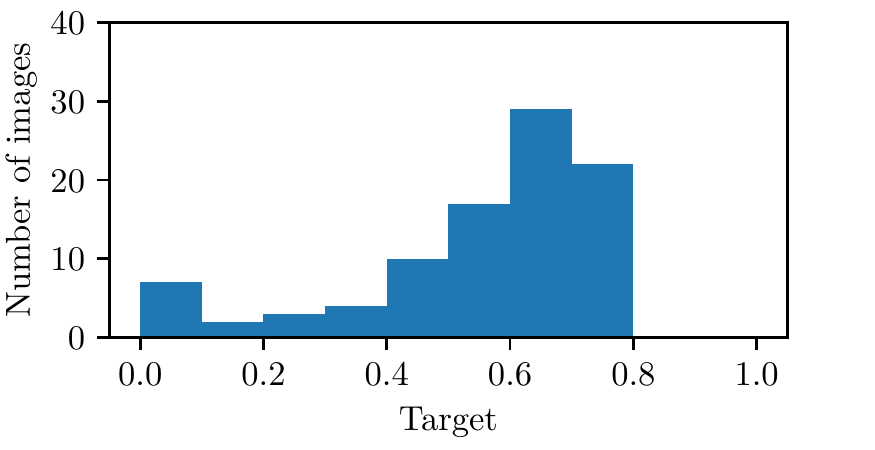}}
    \caption{Distribution of the tubulin dataset.}
\label{fig:data:tubulin:distribution}
\end{figure}

This dataset also contains 94 images, but of a different protein: the tubulin.
Each image was obtained by the imaging of the cytoskelettal protein tubulin stained with a primary antibody recognizing alpha-tubulin (Mouse-anti-alpha-Tubulin, Sigma-Aldrich, 1:500) and a secondary antibody tagged with the fluorophore Alexa594 (Goat-anti-Mouse-Alexa594, Molecular Probes, 1:250). STED imaging of the fluorophore Alexa594 was performed on an Abberior 4-colors STED microscope with  40 MHz pulsed lasers at 561~nm and 775~nm for the excitation and depletion respectively.
Figure~\ref{fig:data:tubulin:distribution} shows the distribution of the dataset across quality scores. Images of this protein have never been seen by the network before.

\subsection{Benchmark}

The experiment is performed by 11 experts.
Each expert performs the experiment for both datasets.
None of these experts were involved in the gathering of the data.
The neural network is compared against a random system that predicts a score by sampling it uniformly from the training labels. Therefore, it predicts scores based on the training distribution (Figure~\ref{fig:data:actin_train_distribution}). Note that for similar, non-uniform, training and testing distributions, such random system is expected to perform well in high density regions.
This experiment should allow us to assess that the proposed neural network system is not limited to learning the training distribution.

\subsection{Results}

\begin{table*}
\caption{Confusion performance of the network given the target quality.}
\label{tab:confusion_bins}
\centering
\begin{tabular}{l|cccc}
\hline
    & \multicolumn{2}{c}{Actin}   & \multicolumn{2}{c}{Tubulin}   \\
Target & Network & Random & Network & Random \\
\hline
 0.00 - 0.20 & $\bm{0.63 \pm 0.25}$ & $0.26 \pm 0.15$ & $\bm{0.58 \pm 0.25}$ & $0.04 \pm 0.09$  \\
 0.20 - 0.40 & $\bm{0.61 \pm 0.29}$ & $0.46 \pm 0.22$ & $\bm{0.77 \pm 0.20}$ & $0.44 \pm 0.23$  \\
 0.40 - 0.60 & $\bm{0.76 \pm 0.10}$ & $\bm{0.76 \pm 0.17}$ & $\bm{0.80 \pm 0.12}$ & $0.44 \pm 0.15$  \\
 0.60 - 0.80 & $\bm{0.84 \pm 0.11}$ & $0.70 \pm 0.09$ & $0.79 \pm 0.20$ & $\bm{0.85 \pm 0.13}$  \\
 0.80 - 1.00 & $\bm{0.68 \pm 0.20}$ & $0.48 \pm 0.37$ & -               & -                \\
\hline
\end{tabular}
\end{table*}

\begin{table*}
\caption{Domination performance of the network given the target quality.}
\label{tab:domination_bins}
\centering
\begin{tabular}{l|cccc}
\hline
    & \multicolumn{2}{c}{Actin}   & \multicolumn{2}{c}{Tubulin}   \\
Target & Network & Random & Network & Random \\
\hline
 0.00 - 0.20 & $\bm{0.28 \pm 0.21}$ & $0.10 \pm 0.09$ & $\bm{0.33 \pm 0.23}$ & $0.02 \pm 0.04$  \\
 0.20 - 0.40 & $\bm{0.32 \pm 0.22}$ & $0.23 \pm 0.11$ & $\bm{0.51 \pm 0.26}$ & $0.16 \pm 0.14$  \\
 0.40 - 0.60 & $\bm{0.48 \pm 0.14}$ & $\bm{0.48 \pm 0.15}$ & $\bm{0.57 \pm 0.18}$ & $0.20 \pm 0.08$  \\
 0.60 - 0.80 & $\bm{0.40 \pm 0.12}$ & $0.30 \pm 0.07$ & $\bm{0.61 \pm 0.18}$ & $0.46 \pm 0.13$  \\
 0.80 - 1.00 & $\bm{0.61 \pm 0.24}$ & $0.25 \pm 0.20$ & -               & -                \\
\hline
\end{tabular}
\end{table*}

Tables~\ref{tab:confusion_bins} and~\ref{tab:domination_bins} show these performance measures per bin of quality scores, averaged over testers, with one standard deviation. More specifically, the performance measures are calculated \emph{for each tester} and their scores are then averaged.

We observe that the proposed network approach beats the random baseline in almost all target quality bins, on both datasets, and regarding both measures.
Note that not enough data are currently available in order to make these results statistically significant and further experiments would be required to this extent.

We also observe that the confusion performance appears to be following the data distribution for the random baseline. This was to be expected since this approach predicts based on the training data prior.
The low probability of sampling a low quality score from the training data given the distribution (Figure~\ref{fig:data:actin_train_distribution}) also explains the poor performance of the random technique in low targets. The confusion performance of the proposed network approach also decreases for the lower target quality bin, such that the network also seems affected by the lack of low quality samples. However, it outperforms the random strategy, hence exhibiting a capacity of generalization. Note that the increasing difficulty for lower targets could also be explained the difficulty to train a sigmoid output toward value close to~0. A different network design might help to overcome this situation. Despite this weakness, the proposed approach beats the baseline.

Surprisingly, the network obtains a high domination performance on the tubulin protein dataset. Recall that images of this particular protein have never been seen by the network. In other words, the network predicts scores that often appear to be even better than the true targets, from the eye of a tester. This is a very interesting situation that raises questions regarding the noise inherent to human labelling as well as the human perception of subtle concepts such as \emph{quality}.

\begin{figure}[t]
    \centering
    \begin{subfigure}[b]{0.47\textwidth}
        \centering
        \includegraphics[width=\textwidth]{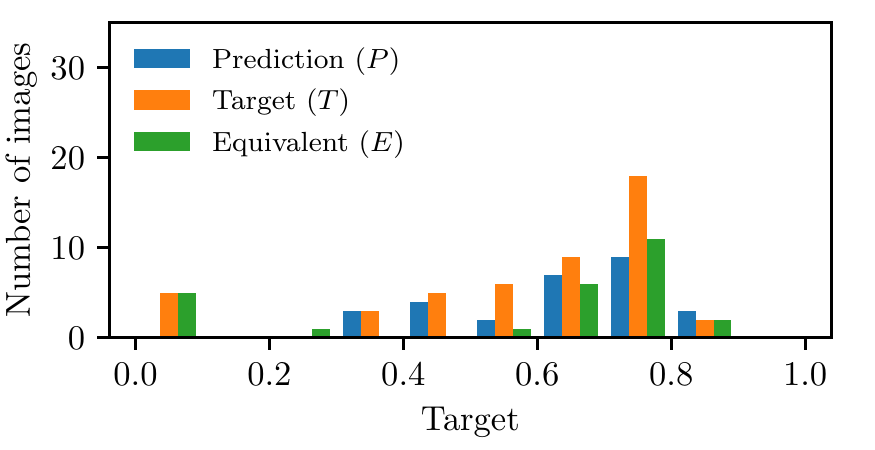}
        \caption{Network}
    \end{subfigure}
    
    \begin{subfigure}[b]{0.47\textwidth}
        \centering
        \includegraphics[width=\textwidth]{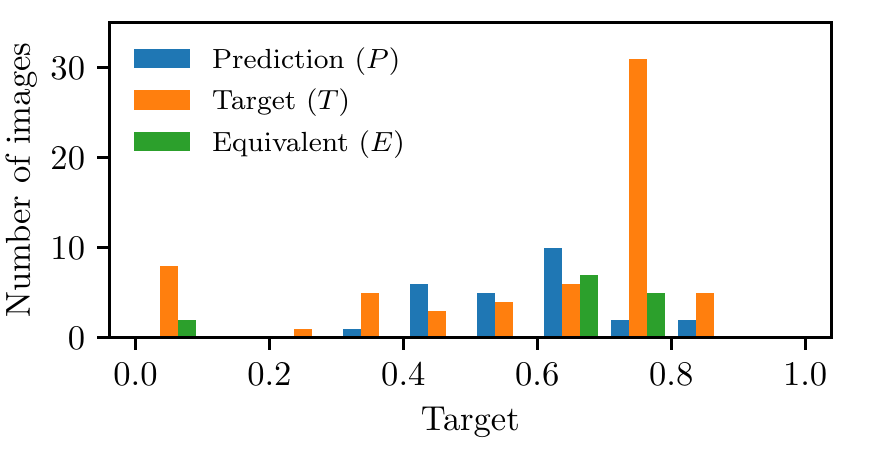}
        \caption{Random}
        \label{fig:results:action:random}
    \end{subfigure}
    \caption{Distributions of effective tester choices per bin of target quality score on the actin dataset.}
\label{fig:results:actin}
\end{figure}

\begin{figure}
    \centering
    \begin{subfigure}[b]{0.47\textwidth}
        \centering
        \includegraphics[width=\textwidth]{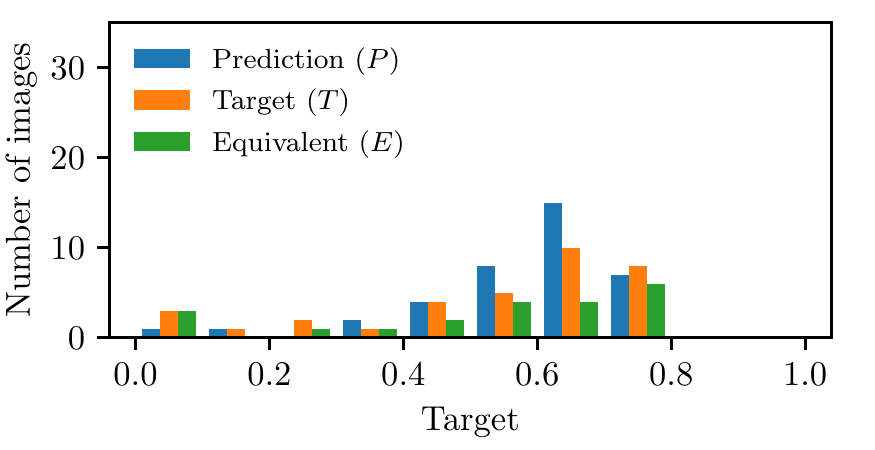}
        \caption{Network}
        \label{fig:results:tubulin:network}
    \end{subfigure}
    
    \begin{subfigure}[b]{0.47\textwidth}
        \centering
        \includegraphics[width=\textwidth]{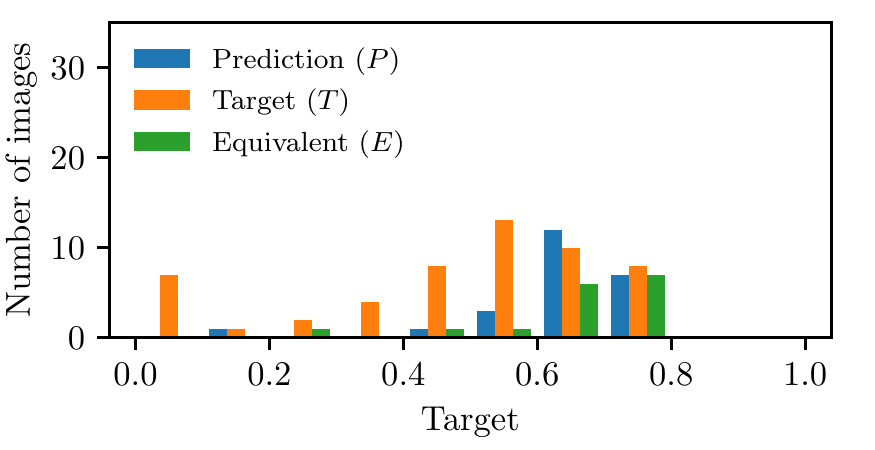}
        \caption{Random}
        \label{fig:results:tubulin:random}
    \end{subfigure}
    \caption{Distributions of effective tester choices per bin of target quality score on the tubulin dataset.}
\label{fig:results:tubulin}
\end{figure}

In order to investigate the contribution of explicit and implicit confusion in the confusion performance score, we visualize the choices of a tester who is close to the means given in Table~\ref{tab:confusion_bins}. Figures~\ref{fig:results:actin} and~\ref{fig:results:tubulin} show
the number of images where the tester picked the target ($T$), the prediction ($P$), or marked them as equivalent ($E$), per bin of target quality score.

Similarly to the confusion results given by Table~\ref{tab:confusion_bins}, we observe that the random technique fails for target quality scores below~0.20.
Both figures show that the explicit confusion (more images marked as equivalent) of the random approach is mostly contained in bins where most of the training distribution (Figure~\ref{fig:results:actin}) density lies.

Another interesting fact is that the random prediction appears to be chosen quite often in the quality bin 0.40 -- 0.60 of the actin dataset (Figure~\ref{fig:results:action:random}), which is not of highest density.
This may be due to a tendency of testers to be more optimistic in their rating of actin images compared with the training expert. This should be investigated further.

As reported by Table~\ref{tab:domination_bins}, we finally observe the network domination phenomena on tubulin images (Figure~\ref{fig:results:tubulin:network}), where network predictions are selected more often than targets.
These results show that, unlike the random technique (Figure~\ref{fig:results:tubulin:random}), the proposed network approach has a generalization capability.

%%%%%%%%%%%%%%%%%%%%%%%%%%
%%%%%%%%%%%%%%%%%%%%%%%%%%
%%       OPENINGS       %%
%%%%%%%%%%%%%%%%%%%%%%%%%%
%%%%%%%%%%%%%%%%%%%%%%%%%%
\section{Openings}

% recap of paper
This paper tackles the problem of learning to evaluate the quality of STED images, which can be formalized as a regression problem in the space of images. We address this using a neural network to model the quality function. A dataset was built especially for this task. The system is evaluated using two datasets: a split of the initial dataset and images of a different protein that the network has never seen. The system is compared against a random technique that predicts the quality solely based upon the prior distribution of the training data. A user study involving four human experts is performed to assess the capability of the proposed approach. Results from the user study show that the network has the potential to mimic a human from an expert perspective.

% raised questions
The obtained results raise several questions. For example, how can predictions of the proposed network become better than actual scores given by a human expert? Could we use the resulting network to help understanding the quality scoring process by a human expert?
More specifically, given that the quality function is driven by the appearance of specific structures, could the resulting network be able to detect these structures?

% highlight contributions
This application is a first step toward the automatization of analysis and optimization tasks for neuroscientists working with high-end microscopy settings.
Thanks to the early promising results, the system has been fully deployed on a STED setting and is currently being used in a control loop for optimizing imaging parameters.
These tools have the potential to help users in taking full advantage of these systems, which could facilitate the adoption of this powerful technique.

%%%%%%%%%%%%%%%%%%%%%%%%%%
%%%%%%%%%%%%%%%%%%%%%%%%%%
%%    ACKNOLEDGMENTS    %%
%%%%%%%%%%%%%%%%%%%%%%%%%%
%%%%%%%%%%%%%%%%%%%%%%%%%%
% \section{Acknowledgments}

\bibliography{references}

\begin{thebibliography}{}

\bibitem[\protect\citeauthoryear{Banterle \bgroup et al\mbox.\egroup
  }{2013}]{Banterle2013}
Banterle, N.; Bui, K.~H.; Lemke, E.~A.; and Beck, M.
\newblock 2013.
\newblock Fourier ring correlation as a resolution criterion for
  super-resolution microscopy.
\newblock {\em Journal of structural biology} 183(3):363--367.

\bibitem[\protect\citeauthoryear{Durisic, Cuervo, and
  Lakadamyali}{2014}]{Durisic2014}
Durisic, N.; Cuervo, L.~L.; and Lakadamyali, M.
\newblock 2014.
\newblock Quantitative super-resolution microscopy: pitfalls and strategies for
  image analysis.
\newblock {\em Current opinion in chemical biology} 20:22--28.

\bibitem[\protect\citeauthoryear{Gardner \bgroup et al\mbox.\egroup
  }{2017}]{Gardner2017}
Gardner, M.-A.; Sunkavalli, K.; Yumer, E.; Shen, X.; Gambaretto, E.; Gagn\'e,
  C.; and Lalonde, J.-F.
\newblock 2017.
\newblock Learning to predict indoor illumination from a single image.
\newblock {\em ACM Transactions on Graphics (SIGGRAPH Asia)} 9(4).

\bibitem[\protect\citeauthoryear{Hell and Wichmann}{1994}]{Hell1994}
Hell, S.~W., and Wichmann, J.
\newblock 1994.
\newblock Breaking the diffraction resolution limit by stimulated emission:
  stimulated-emission-depletion fluorescence microscopy.
\newblock {\em Optics letters} 19(11):780--782.

\bibitem[\protect\citeauthoryear{Klar \bgroup et al\mbox.\egroup
  }{2000}]{Klar2000}
Klar, T.~A.; Jakobs, S.; Dyba, M.; Egner, A.; and Hell, S.~W.
\newblock 2000.
\newblock Fluorescence microscopy with diffraction resolution barrier broken by
  stimulated emission.
\newblock {\em Proceedings of the National Academy of Sciences}
  97(15):8206--8210.

\bibitem[\protect\citeauthoryear{Kraus \bgroup et al\mbox.\egroup
  }{2017}]{Kraus2017}
Kraus, O.~Z.; Grys, B.~T.; Ba, J.; Chong, Y.; Frey, B.~J.; Boone, C.; and
  Andrews, B.~J.
\newblock 2017.
\newblock Automated analysis of high-content microscopy data with deep
  learning.
\newblock {\em Molecular Systems Biology} 13(4):924.

\bibitem[\protect\citeauthoryear{Li \bgroup et al\mbox.\egroup }{2017}]{Li2017}
Li, R.; Zeng, T.; Peng, H.; and Ji, S.
\newblock 2017.
\newblock Deep learning segmentation of optical microscopy images improves 3-d
  neuron reconstruction.
\newblock {\em IEEE Transactions on Medical Imaging} 36(7):1533--1541.

\bibitem[\protect\citeauthoryear{Li, Bovik, and Wu}{2011}]{Li2011}
Li, C.; Bovik, A.~C.; and Wu, X.
\newblock 2011.
\newblock Blind image quality assessment using a general regression neural
  network.
\newblock {\em IEEE Transactions on Neural Networks} 22(5):793--799.

\bibitem[\protect\citeauthoryear{Merino \bgroup et al\mbox.\egroup
  }{2017}]{Merino2017}
Merino, D.; Mallabiabarrena, A.; Andilla, J.; Artigas, D.; Zimmermann, T.; and
  Loza-Alvarez, P.
\newblock 2017.
\newblock Sted imaging performance estimation by means of fourier transform
  analysis.
\newblock {\em Biomedical Optics Express} 8(5):2472--2482.

\bibitem[\protect\citeauthoryear{Sahl, Hell, and Jakobs}{2017}]{Sahl2017}
Sahl, S.~J.; Hell, S.~W.; and Jakobs, S.
\newblock 2017.
\newblock Fluorescence nanoscopy in cell biology.
\newblock {\em Nature reviews. Molecular cell biology}.

\bibitem[\protect\citeauthoryear{Vohs \bgroup et al\mbox.\egroup
  }{2008}]{Vohs2008}
Vohs, K.~D.; Baumeister, R.~F.; Schmeichel, B.~J.; Twenge, J.~M.; Nelson,
  N.~M.; and Tice, D.~M.
\newblock 2008.
\newblock Making choices impairs subsequent self-control: A limited-resource
  account of decision making, self-regulation, and active initiative.
\newblock {\em Journal of Personality and Social Psychology} 94(5):883--898.

\bibitem[\protect\citeauthoryear{Willig \bgroup et al\mbox.\egroup
  }{2006}]{Willig2006}
Willig, K.~I.; Kellner, R.; Medda, R.; Hein, B.; Jakobs, S.; and Hell, S.~W.
\newblock 2006.
\newblock Nanoscale resolution in gfp-based microscopy.
\newblock {\em Nature Methods} 3(9):721--723.

\end{thebibliography}
\bibliographystyle{aaai}

\end{document}